\documentclass[runningheads]{llncs}

\usepackage{eccv}

\usepackage{eccvabbrv}

\usepackage{graphicx}
\usepackage{booktabs}

\usepackage[accsupp]{axessibility}  

\usepackage{hyperref}

\usepackage{orcidlink}

\usepackage[table,xcdraw,dvipsnames]{xcolor}
\usepackage{threeparttable}
\usepackage{makecell}
\usepackage{xspace}
\usepackage{amsmath}
\usepackage{multirow}
\usepackage{xurl}
\usepackage{caption}
\usepackage{pifont}
\usepackage{array}
\newcommand{\TODO}[1]{\textbf{\color{red}[TODO: #1]}}

\newcommand{\cmark}{\ding{51}}  
\newcommand{\xmark}{\ding{55}}  

\newcommand{\methodname}{RelGraphOV}
\newcommand{\method}{{\methodname}\xspace}

\colorlet{colorFst}{Green!25}       %
\colorlet{colorSnd}{SpringGreen!50} %
\colorlet{colorTrd}{Yellow!35}      %

\newcommand{\fs}{\cellcolor{colorFst}\textbf}
\newcommand{\nd}{\cellcolor{colorSnd}}
\newcommand{\rd}{\cellcolor{colorTrd}}

\renewcommand{\TODO}[1]{}

\begin{document}

\title{Beyond Isolated Objects: Relationship-aware Open Vocabulary Scene Understanding via 3D Scene Graph Analysis} 

\titlerunning{RelGraphOV}

\author{Xianhao Chen\inst{1}$^{\ast}$\orcidlink{0009-0004-7451-1871} \and
Jiarui Hu\inst{1}$^{\ast}$\orcidlink{0009-0006-9563-8956} \and
Yuanbo Yang\inst{2}\orcidlink{0009-0006-3442-1024} \and
Xiyu Zhang\inst{1}\orcidlink{0009-0001-6001-133X} \and
Tengyue Wang\inst{2}\orcidlink{0009-0007-9395-0417} \and
Hujun Bao\inst{1}\orcidlink{0000-0002-2662-0334} \and
Guofeng Zhang\inst{1}\orcidlink{0000-0001-5661-8430} \and
Zhaopeng Cui\inst{1}$^\dag$\orcidlink{0000-0002-7130-439X}}

\authorrunning{Chen et al.}

\institute{State Key Lab of CAD\&CG, Zhejiang University, China\\ \and
Zhejiang University, China}

\maketitle
\begingroup
\renewcommand{\thefootnote}{}
\footnotetext[0]{$^\ast$ Equal contribution. $^\dag$ Corresponding author.}
\endgroup

\begin{abstract}
Open-vocabulary 3D scene understanding aims to segment 3D scenes beyond predefined categories by transferring semantic knowledge from vision-language models. Existing methods have advanced this task by lifting language-aligned 2D features into 3D, yet they often rely on context-independent semantic representations, leaving object relationships underexplored for contextual refinement. We propose \textit{\method{}}, a relationship-aware framework that uses 3D scene graphs to enhance open-vocabulary 3D understanding. Our method constructs relational scene graphs from multi-view observations by leveraging vision-language reasoning to infer object relationships and prune geometrically implausible connections, without manual relationship annotations. To aggregate relational context while avoiding feature interference, we introduce an Adaptive Gated Dual-Stream Contextual GAT that separates dense geometric features and semantic CLIP embeddings, performs edge-guided message passing, and adaptively fuses complementary semantics. A hierarchical contrastive objective further promotes instance-level consistency and category-level discrimination. Experiments on ScanNetV2, ScanNet200, ScanNet$++$, and Replica demonstrate strong performance and generalization ability. Project Page: \href{https://cxavireh.github.io/relgraphov-projectpage}{cxavireh.github.io/relgraphov-projectpage}

\end{abstract}

\section{Introduction}
\label{sec:intro}

\begin{figure*}[t!]
  \centering
  \includegraphics[width=0.98\textwidth]{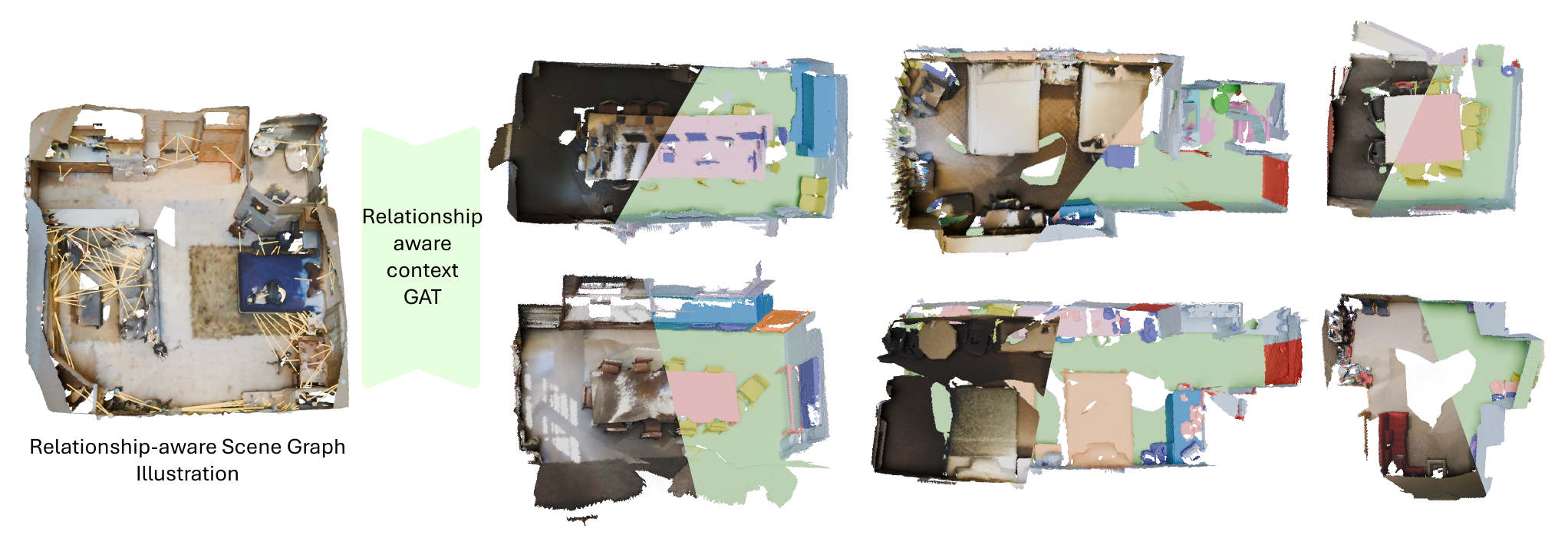}
  \caption{\textbf{\method{}.} 
  We propose a relationship-aware open-vocabulary 3D scene understanding framework. 
  Left: Illustration of encoding a 3D scene into a relationship-aware scene graph to capture contextual relationships. 
  Right: Qualitative open-vocabulary segmentation results produced by our method.}
  \label{fig:teaser}
\end{figure*}

Open-vocabulary 3D scene understanding primarily aims to perform dense semantic segmentation of 3D environments, extending object recognition beyond a fixed set of predefined categories to enable scalable perception for applications such as robotic navigation, spatial reasoning, and augmented reality. Inspired by vision-language models such as CLIP~\cite{clip}, recent works transfer semantic knowledge learned from image-text data into 3D representations. Existing open-vocabulary 3D semantic segmentation approaches generally fall into two paradigms. The first projects \emph{dense vision-language features} from 2D models into 3D space~\cite{openscene,cuao3d}, which provides strong spatial robustness under occlusion or incomplete observations. The second associates \emph{instance-level CLIP embeddings} with 3D object proposals~\cite{openmask3d,maskclustering,mpec}, enabling stronger semantic generalization to long-tail categories. Despite their complementary strengths, both paradigms primarily rely on context-independent semantic representations, leaving structured object relationships underexplored for semantic refinement. For instance, distinguishing a ``curtain'' from a ``shower curtain'' is notoriously difficult when relying exclusively on isolated object appearances. However, if the model is aware of the surrounding context, such as the object's spatial relationship to a bathtub or a toilet, this semantic ambiguity can be naturally resolved.

A natural representation for modeling such contextual information is the \emph{3D scene graph}, which represents objects as nodes and their relationships as edges. Existing relational 3D methods have studied scene graph prediction~\cite{wald2020learning,scenegraphfusion,vlsat,lang3dsg,open3dsg}, relationship querying~\cite{conceptgraph,relationfield}, referred object grounding~\cite{contextaware3dgrounding}, and relation-aware instance segmentation~\cite{relation3d}. However, they do not directly address how inferred 3D semantic relationships can serve as intermediate contextual priors to refine open-vocabulary semantic features for dense 3D segmentation. Introducing scene graphs into this setting poses two major challenges. First, \textit{data preparation} becomes difficult because open-vocabulary settings significantly increase the diversity of objects and relationships, while large-scale relationship annotations are rarely available. Second, aggregating heterogeneous representations on the graph leads to a fundamental \textit{feature granularity dilemma}. Dense features provide strong spatial robustness but limited semantic generalization, whereas CLIP-based instance features offer richer semantics but are sensitive to partial observations. Although these representations appear complementary, naively combining them within graph neural networks often causes severe feature interference, where geometric noise contaminates the semantic embedding space during message passing.

To address these challenges, we propose \textit{\method{}} (see Fig.~\ref{fig:teaser}), a relationship aware open-vocabulary 3D scene understanding framework that explicitly incorporates contextual reasoning via 3D scene graphs. Our framework first constructs a relational scene graph from multi-view observations and then performs contextual feature learning on the graph to enhance open-vocabulary representations. Specifically, to alleviate the lack of relationship annotations, we design a scene graph construction pipeline that leverages multi-view observations and vision-language reasoning to infer object relationships and filter geometrically implausible connections. This process enables labor-free and efficient graph construction without manual relationship labeling and provides reliable relational priors for contextual feature aggregation. Different from prior relational 3D methods that mainly target scene graph construction, querying, grounding, or instance-level reasoning, our approach uses the constructed graph as a relational structure for contextual feature learning in open-vocabulary 3D segmentation.

To further address the feature granularity dilemma and prevent feature interference during graph-based aggregation, we propose an Adaptive Gated Dual-Stream Contextual GAT. Instead of directly mixing heterogeneous representations, our architecture decouples feature propagation into two streams. The main stream aggregates contextual information using dense LSeg features along scene graph edges, while the auxiliary stream independently preserves high-level semantic embeddings extracted from CLIP. A learnable gating mechanism dynamically injects complementary semantic cues from the auxiliary stream during message passing, enriching node representations while preventing cross-modal contamination. In addition, we introduce a hierarchical contrastive learning strategy that jointly enforces instance-level consistency and category-level discrimination, effectively mitigating semantic drift in open-vocabulary settings.

We evaluate our method on ScanNetV2~\cite{scannet}, ScanNet200~\cite{scannet200}, and further conduct strict cross dataset zero-shot evaluation on ScanNet$++$\cite{scannetpp} and Replica~\cite{replica}. Experimental results demonstrate that \textit{\method{}} consistently outperforms existing methods and achieves state-of-the-art performance in open-vocabulary 3D scene understanding.

To summarize, our contributions are as follows:

\begin{itemize}
\item We propose \textit{\method{}}, a relationship-aware open-vocabulary 3D scene understanding framework that incorporates contextual reasoning via automatically constructed 3D scene graphs.

\item We introduce a multi-view reasoning-based scene graph construction pipeline that enables scalable generation of object relationships without manual annotation.
\item We propose an Adaptive Gated Dual-Stream Contextual GAT with hierarchical contrastive learning to address the feature granularity dilemma and mitigate feature interference during graph-based context aggregation.
\item Extensive experiments on ScanNetV2, ScanNet200, ScanNet$++$, and Replica demonstrate state-of-the-art performance and strong zero-shot generalization.

\end{itemize}

\section{Related Work}
\label{sec:relatedwork}

\noindent{\textbf{2D Visual Foundation Models.}}
Recent advances in large-scale pre-training have established Visual Foundation Models (VFMs) as the cornerstone for 2D visual perception. CLIP~\cite{clip} aligns images and text through contrastive pre-training, enabling zero-shot recognition, while DINO~\cite{dino, dinov2} learn transferable dense visual features. These models have been widely used for segmentation~\cite{sam, sam2, maskattentionmask, oneformer, seeaao}, detection~\cite{groundingdino}, captioning~\cite{osprey, dam}, and open-vocabulary recognition~\cite{silc, catseg}. In 3D scene understanding, they serve as semantic priors for transferring language-aligned knowledge from 2D observations to 3D representations.

\noindent{\textbf{Open-Vocabulary 3D Scene Understanding.}}
Conventional 3D semantic segmentation methods~\cite{mink,mix3d} are typically trained within predefined label spaces and require dense annotations, limiting their generalization to novel categories. Recent open-vocabulary 3D methods address this limitation by transferring 2D vision-language features into 3D scenes. OpenScene~\cite{openscene} pioneered this direction by distilling dense language-aligned 2D features, such as LSeg~\cite{lseg} and OpenSeg~\cite{openseg}, into 3D point clouds. Subsequent methods~\cite{regionplc, ov3d, dma, cuao3d, mosaic3d, opennerf} improve 2D--3D feature alignment, multi-view feature fusion, and open-vocabulary querying. Instance-level methods such as OpenMask3D~\cite{openmask3d} and related approaches~\cite{mpec} further associate object proposals with CLIP-like features for object-level recognition. While effective, these methods often treat objects in a scene as isolated entities, leaving structured object relations underexplored for context-aware semantic refinement.

\noindent{\textbf{Relational 3D Scene Understanding.}}
Object relations have been explored in 3D scene graph prediction and relation-aware perception. Existing 3D scene graph methods predict object nodes and semantic relationships from reconstructed scenes or RGB-D sequences~\cite{wald2020learning, scenegraphfusion, conceptgraph}, with recent works further incorporating visual-linguistic supervision, language-based pre-training, or open-vocabulary relationship prediction~\cite{vlsat,lang3dsg,open3dsg}. Beyond scene graph prediction, relations have also been used for referred entity grounding~\cite{contextaware3dgrounding}, relational querying in radiance fields~\cite{relationfield}, and closed-set instance segmentation via superpoint-level geometric relations~\cite{relation3d}. These studies show the value of explicit object relations for 3D scene understanding. Our work studies a complementary use of such relations in open-vocabulary 3D semantic segmentation: the constructed scene graph is used as a contextual structure for aggregating features among object proposals, so that language-aligned 3D features can be refined before dense point-level labeling.

\section{Method}
\label{sec:methods}

\begin{figure*}[t!]
  \centering
  \includegraphics[width=0.98\textwidth]{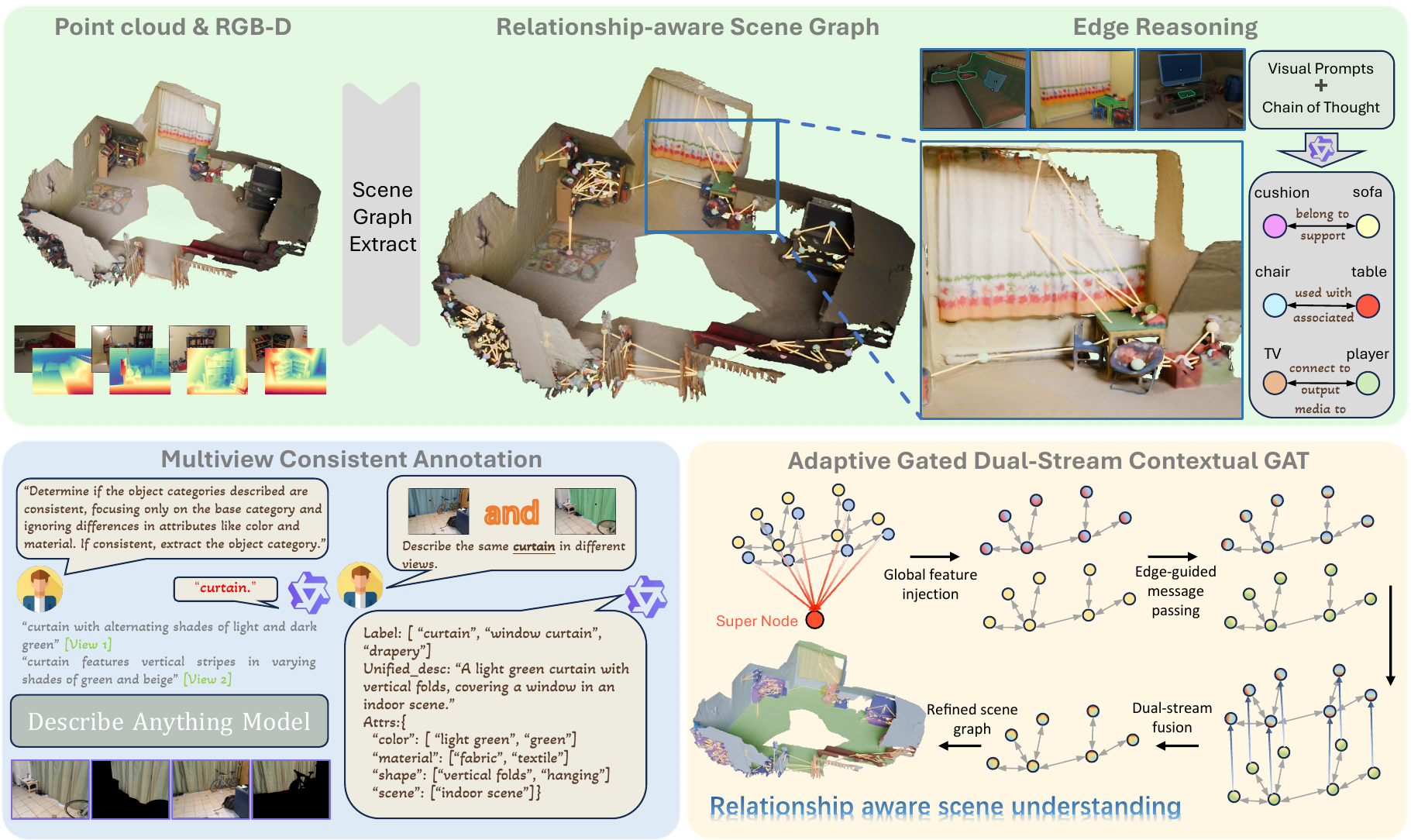}
  \caption{\textbf{Overall Pipeline of \method{}.} (1) \textit{Graph Construction}: We build a relationship-aware 3D scene graph via multi-view VLM reasoning. (2) \textit{Annotation}: A VLM chain-of-thought generates rich node semantics. (3) \textit{Adaptive Gated Dual-Stream GAT}: To prevent feature interference, we decouple geometric (main) and semantic (auxiliary) feature propagation. Both undergo edge-guided message passing, while a learnable gate dynamically injects auxiliary semantics into the main stream. (4) \textit{Training}: Hierarchical contrastive and dual-alignment losses optimize the network, effectively mitigating semantic drift and catastrophic forgetting.}
  \label{fig:pipeline}
\end{figure*}

\noindent{\textbf{Problem Statement.}}
As illustrated in Fig. \ref{fig:pipeline}, 3D open-vocabulary scene understanding aims to segment 3D scenes with free-form textual descriptions, going beyond a fixed set of predefined categories while supporting natural-language queries. The system processes each point in the scene by aligning its 3D features with semantic embeddings derived from natural language. This enables accurate point-level labeling and retrieval of semantically relevant regions based on textual descriptions.

\noindent\textbf{Overview.}
As illustrated in Fig. \ref{fig:pipeline}, the input to our system consists of RGB-D sequences and point clouds, where the point clouds can be obtained either from SLAM algorithms~\cite{orbslam3, cpslam, cgslam} or dense 3D reconstruction~\cite{vggt, atlasgs}. The first step (Sec. \ref{subsec:Scene graph construction}) is to abstract the scene into a relationship-aware 3D scene graph, where keyframes are extracted using a unified viewpoint scoring mechanism, node features are initialized, and semantic relationships for each edge are reasoned by a Vision-Language Model (VLM). In the second step (Sec. \ref{subsec:VLM data engine}), we generate node annotations, embedding rich semantic information by leveraging the powerful captioning capabilities of a VLM. The third step (Sec. \ref{subsec:GAT}) involves employing an Adaptive Gated Dual-Stream Contextual GAT, which decouples multi-modal feature propagation and utilizes edge-feature-guided message passing to optimize node semantics without suffering from feature interference. Finally, we design a hierarchical contrastive loss with auxiliary supervision (Sec. \ref{subsec:training}) to train the model based on its dual-stream characteristics and the 3D scene graph structure.

\subsection{Relationship-Aware Scene Graph Construction}
\label{subsec:Scene graph construction}

To effectively utilize the environmental context within the scene for improved scene understanding, it is crucial to construct a relationship-aware open vocabulary 3D scene graph. First, following MaskClustering~\cite{maskclustering}, we aggregate 2D masks from RGB-D sequences to obtain 3D class-agnostic object proposals, represented as point cloud clusters \(\mathcal{M} = \{ M_1, M_2, \dots, M_N \} \). We treat such proposal generation and feature extraction as interchangeable upstream modules rather than our focus, so \method{} inherits their proposal quality while remaining agnostic to their specific choice. This modular design also allows our relational refinement to directly benefit from increasingly powerful proposal generators and vision-language backbones without architectural changes. Taking these 3D proposals as input, the scene graph construction process comprises four key steps.

\noindent\textbf{Node Definition, Keyframe Extraction, and Pruning.}
We treat each proposal \(M_i\) as a candidate node \(v_i\) and filter out small, meaningless instances based on the number of points. For each node \(v_i\), we calculate its 3D axis-aligned bounding box \(B_i\) and centroid \(C_i\). At this stage, to obtain comprehensive visual observations for each node, we extract representative keyframes from the original RGB-D sequence. Specifically, by projecting the 3D bounding box onto the 2D frames, we evaluate each candidate frame \(I_k\) using a universal composite scoring function \(S(I_k, W)\):
\begin{equation}
\begin{split}
S(I_k, W) &= W \cdot \text{S}, \quad W = \begin{bmatrix} w_{\text{comp}}, w_{\text{area}}, w_{\text{center}}, w_{\text{sharp}} \end{bmatrix}, \\
\text{S} &= \begin{bmatrix} s_{\text{comp}}, s_{\text{area}}, s_{\text{center}}, s_{\text{sharp}} \end{bmatrix}^T,
\end{split}
\label{eq:scoring}
\end{equation}
\noindent where \(s_{\text{comp}}, s_{\text{area}}, s_{\text{center}}, s_{\text{sharp}} \in (0, 1)\) represent the node completeness, projected area, center score, and motion blur sharpness, respectively. \(W\) is a customizable weight vector tailored to different selection objectives. Using this mechanism, we select keyframes offering optimal visibility and centrality for each object. Following this, we perform structural pruning. Nodes representing large, non-interactive background structures, such as floors, walls, and ceilings, are identified and removed based on specific geometric attributes, including volume and surface normal direction. This ensures the graph focuses exclusively on core interactive objects and avoids redundant connections.

\noindent\textbf{Node Feature Initialization.}
With the keyframes extracted, we proceed to acquire the multi-modal features for each node. We project the 3D point cloud of each node onto its corresponding 2D keyframes to extract pixel-aligned dense features using LSeg~\cite{lseg} and instance-level global features using CLIP~\cite{clip}. The extracted 2D features are pooled to form the base 3D LSeg semantic feature \(f_{l}\) and instance semantic CLIP feature \(f_{c}\) for each node. Consequently, the comprehensive multimodal feature representation for each node is formulated as \(\mathbf{F}_v = \{f_l, f_c\}\), which serves as the decoupled inputs for our dual-stream architecture.

\noindent\textbf{Edge Initialization.}
We initialize the graph's connectivity \( \mathbf{E}_{geom}\) based on spatial proximity. An edge \(e_{ij}\) is established either if the nodes are among the three nearest neighbors based on Euclidean distance between centroids, or if their 3D bounding boxes are adjacent or overlap. This constructs a dense and geometrically coherent candidate graph.

\noindent\textbf{VLM-based Edge Refinement and Encoding.}
The final step is the core of our relational construction, where we refine each candidate edge \(e_{ij} \in \mathbf{E}_{geom}\) using a VLM. Specifically, we first select a representative frame \(I_{v_{i}v_{j}}\) that clearly captures both nodes by jointly applying the unified scoring criterion (Eq.~\ref{eq:scoring}) to their projected regions. Given this optimal viewpoint, we draw inspiration from Set-of-Mark~\cite{SoM} to visually prompt the VLM agent, employing a chain-of-thought~\cite{cot} reasoning process to explicitly describe the inter-object relations. Based on the VLM's response, any candidate edge returning a ``\textit{none}'' relationship is immediately pruned. The subset of spatial connections successfully validated by the VLM constitutes our refined edge set, denoted as \( \mathbf{E}_{refine} \). For these valid edges, the VLM generates two asymmetric descriptions (e.g., ``\textit{person sits on chair}'' and ``\textit{chair supports person}''). To formally encode the edge feature \( \mathbf{F}_e\), we parse these descriptions to extract Subject-Verb-Object (SVO) triplets. We then feed the subject, verb, and object independently into the CLIP text encoder and concatenate the resulting three feature vectors to construct a highly-structured edge representation. This process yields the refined relationship-aware scene graph:
\begin{equation}
    \mathbf{SG} = (\mathbf{V}, \mathbf{E}_{refine}, \mathbf{F}_v, \mathbf{F}_e).
\end{equation}

\subsection{VLM-based Scene Graph Annotation Engine}
\label{subsec:VLM data engine}
The constructed graph requires accurate node-level annotations for supervision. However, the same 3D object viewed from different 2D perspectives may yield conflicting descriptions. We design an automated engine to adjudicate this multi-view inconsistency.

\noindent\textbf{Multi-View Selection and View-Specific Captioning.}
For each node \(v_i\), we project its point cloud onto all RGB-D frames. To rank these frames, we utilize the unified composite scoring function \( S(I_k, W)\) defined in Eq.~\ref{eq:scoring}. By adjusting the weight vector \(W\), we first choose the best viewpoint \(I_1\) emphasizing instance visibility. Then, we select a second viewpoint \(I_2\) by applying a different set of weights emphasizing centrality and sharpness, accompanied by an IoU penalty to ensure spatial dissimilarity. These viewpoints are fed into SAM~\cite{sam} with sampled point prompts to generate object masks \(M_1^{2D}\) and \(M_2^{2D}\), which are subsequently processed by the Describe Anything Model~\cite{dam} to generate initial viewpoint-specific descriptions \(D_1\) and \(D_2\).

\noindent\textbf{VLM Adjudication and Refinement.}
To ensure consistency, a lightweight LLM assesses if the core category recognition in \(D_1\) and \(D_2\) aligns. If consistent, a guided prompt instructs the VLM to refine attributes using both views. Otherwise, an adjudication prompt explicitly informs the VLM of the conflict, directing it to use visual content from both views as the ``final truth'' to determine the correct category. This effectively filters out background noise and yields highly accurate textual annotations. The final textual annotation of node \(v_i\) is encoded by the CLIP text encoder as \(f_{anno}^{i}\), which is used as soft semantic supervision during training.

\subsection{Adaptive Gated Dual-Stream Contextual GAT}
\label{subsec:GAT}

After obtaining the scene graph \(\mathbf{SG}\) and VLM edge features \(\mathbf{F}_e\), our objective is to refine the context-independent features into context-aware representations. However, naively mixing LSeg features \(f_l\) with instance-level CLIP features \(f_c\) during graph aggregation causes severe feature interference. To resolve this, we propose an Adaptive Gated Dual-Stream GAT. To ensure mathematical clarity, we use subscript $m$ to denote the main stream and $a$ to denote the auxiliary stream. The network architecture and refinement process consist of the following carefully decoupled modules.

\noindent\textbf{Modality-Specific Node Initialization.} 
We establish two decoupled initial states to prevent early contamination. The center point coordinates and bounding box dimensions are encoded via a Fourier Feature Encoder and an MLP. These geometric features are concatenated with the dense LSeg features (\(f_l\)) to form the initial state \(h_{m, i}^{(0)}\) for the main stream, and concatenated with the projected CLIP features (\(f_c\)) to form the initial state \(h_{a, i}^{(0)}\) for the auxiliary stream.

\noindent\textbf{Global Context Aggregation.} 
To prevent over-smoothing and expand the receptive field, we dynamically introduce a virtual \textbf{\textit{super node}} \(v_{super}\) connected to all real nodes \(v_i\), forming an augmented graph \(G_{aug}\). Processed by a standard GAT module, the output state of \(v_{super}\) serves as the global context feature \(G_{global}\).

\noindent\textbf{Global Context Injection.} 
During each layer \(l\) of the GAT, we first inject the global context exclusively into the main stream via a Cross-Attention module. Using the main stream node state \(h_{m, i}^{(l)}\) as the Query, and the global feature \(G_{global}\) as both Key and Value, the context-modulated feature is merged with \(h_{m, i}^{(l)}\) through an MLP. This produces a global-aware state \( \tilde{h}_{m, i}^{(l)} \), explicitly bridging local geometry with scene-level understanding. 

\noindent\textbf{Edge-Feature-Guided Message Propagation.}
Following the global injection, both streams perform independent neighborhood aggregation. To emphasize the critical role of relationship semantics, we design an edge-feature-guided propagation mechanism. Specifically, for the main stream, we use the node's global-aware state \( \tilde{h}_{m, i}^{(l)} \) as the \textbf{Query}. For the \textbf{Key}, we concatenate the neighbor node's state \( \tilde{h}_{m, j}^{(l)} \) with the edge feature \(f_{e_{ij}}\). The \textbf{Value} is strictly defined as the neighbor node's state \( \tilde{h}_{m, j}^{(l)} \). By injecting \(f_{e_{ij}}\) into the Key, the cross-attention mechanism adaptively modulates the attention weights, forcing the GNN to amplify information from strongly relevant neighbors and suppress irrelevant ones based solely on their semantic relationships. The auxiliary stream undergoes a similarly independent propagation using its own state \( h_{a, i}^{(l)} \). We denote the intermediate aggregated states after this step as \( \hat{h}_{m, i}^{(l)} \) and \( \hat{h}_{a, i}^{(l)} \).

\noindent\textbf{Adaptive Gated Fusion.}
After the independent message passing, the final step in the layer update is to allow the main stream to dynamically borrow richer, highly generalizable semantics from the auxiliary stream. We achieve this via a context-aware fusion gate:
\begin{equation}
    Gate = \sigma(\text{MLP}([\hat{h}_{m, i}^{(l)} \parallel \hat{h}_{a, i}^{(l)}])),
\end{equation}
\begin{equation}
    h_{m, i}^{(l+1)} = \hat{h}_{m, i}^{(l)} + \alpha \cdot Gate \odot \hat{h}_{a, i}^{(l)},
\end{equation}
where $\alpha$ is a scaling factor. Concurrently, the auxiliary stream securely preserves its pure semantics for the next layer (\(h_{a, i}^{(l+1)} = \hat{h}_{a, i}^{(l)}\)). This unilateral gating mechanism allows autonomous calibration of the main semantic representation without contaminating the auxiliary open-vocabulary priors.

\noindent\textbf{Feature Decoders.} 
After \(L\) layers, the main stream feature \(h_{m, i}^{(L)}\) is processed by an MLP decoder to yield the final context-aware prediction \(F_{out}\). Concurrently, the auxiliary stream employs an independent decoder to output \(F_{aux}\), used exclusively for auxiliary supervision.

\subsection{Dual-Stream Model Training}
\label{subsec:training}
\noindent\textbf{Learning Objective.} 
The training objective of our model is to learn a refinement process that transforms context-independent features \(F_{v}\) into context-aware features \(F_{out}\). Since we use the open-vocabulary annotations generated in Sec. \ref{subsec:VLM data engine} as supervision, a standard contrastive loss would penalize semantically similar categories. We design a Hierarchical Contrastive Loss (\(\mathcal{L}_{hie}\)) for the main stream to solve this issue, along with dual-alignment losses (\(\mathcal{L}_{reg}\) and \(\mathcal{L}_{aux}\)) to prevent catastrophic forgetting.

\noindent\textbf{Loss Design.} 
We design a composite loss function:
\begin{equation}
    \mathcal{L}_{total} = \lambda_{\text{hie}} \cdot \mathcal{L}_{\text{hie}} + \lambda_{\text{reg}} \cdot \mathcal{L}_{\text{reg}} + \lambda_{\text{aux}} \cdot \mathcal{L}_{\text{aux}}.
\end{equation}
The core term is \(\mathcal{L}_{\text{hie}}\),
which encourages the refined node feature \(F_{out}\) to satisfy three objectives: 
(1) self-consistency with its corresponding annotation feature \(f_{anno}^{i}\), 
(2) intra-class consistency among semantically similar instances, and 
(3) inter-class separation from unrelated categories.

To model intra-class consistency, we construct an inferred positive mask \(M_{pos}\), where \(M_{pos}^{ij}=1\) if the cosine similarity between annotation features \(f_{anno}^{i}\) and \(f_{anno}^{j}\) exceeds a threshold \(\tau\). 
To further emphasize self-consistency, we introduce a margin (\(\mathcal{M}\)) to the self-pair similarity:
\begin{equation}
    S_{ij} = \frac{F_{out}^{v_i} \cdot f_{anno}^{j}}{\eta} + \delta_{ij} \cdot \mathcal{M},
\end{equation}
\begin{equation}
\mathcal{L}_{hie}
= -\frac{1}{|\mathbf{V}|} \sum_{i \in \mathbf{V}} \sum_{j \in \mathbf{V}}
\log \frac{M_{\mathrm{pos}}^{ij} \exp(S_{ij})}{\sum_{k \in \mathbf{V}} \exp(S_{ik})},
\end{equation}
where \(\eta\) denotes the temperature parameter and \(\delta_{ij}\) is the Kronecker delta.

In addition, both the regularization loss \(\mathcal{L}_{reg}\) and the auxiliary loss \(\mathcal{L}_{aux}\) follow a shared dual-alignment formulation designed to balance learning new open-vocabulary semantics while preserving modality-specific priors. We define a generalized alignment loss:
\begin{equation}
\mathcal{L}_{align}(F, f_{init}, \gamma)
= \frac{1}{N} \sum_{i=1}^{N}
\left[
\gamma (1 - F^{v_i} \cdot f_{anno}^{i})
+
(1-\gamma)(1 - F^{v_i} \cdot f_{init}^{v_i})
\right],
\end{equation}
where \(f_{init}\) denotes the initial modality feature and \(\gamma\) controls the trade-off between adapting to VLM annotations and preserving the original feature prior.

Accordingly, we define
\(\mathcal{L}_{reg} = \mathcal{L}_{align}(F_{out}, f_l, \gamma_{reg})\) 
to regularize the main stream by aligning it with both the VLM annotation \(f_{anno}\) and the dense feature prior \(f_l\). 
Similarly, 
\(\mathcal{L}_{aux} = \mathcal{L}_{align}(F_{aux}, f_c, \gamma_{aux})\) 
guides the auxiliary stream to learn from the same annotations while remaining anchored to its original CLIP representation \(f_c\). 
This symmetric dual-alignment design enables both streams to absorb contextual supervision while preserving their complementary modality strengths during graph propagation.

\section{Experiments}
\label{sec:experiments}

\subsection{Experimental Setup}
\noindent\textbf{Datasets.}
To evaluate the effectiveness of our proposed method, we conduct experiments on widely used indoor 3D datasets, ScanNetV2~\cite{scannet} and ScanNet200~\cite{scannet200}, along with ScanNet$++$~\cite{scannetpp} and the photorealistic synthetic dataset Replica~\cite{replica}. For ScanNetV2~\cite{scannet}, we follow the official data split (1,201 scenes for training, 312 for validation) and adhere to the standard 20-class evaluation subset to ensure a fair comparison with prior work. ScanNet200~\cite{scannet200} extends the original annotations to a 200-category label space, offering a more fine-grained and challenging benchmark to assess model robustness in open-vocabulary scenarios. We further evaluate cross-dataset zero-shot generalization on ScanNet$++$ and Replica without additional training or fine-tuning, following the standard 100-class benchmark for ScanNet$++$ and the 51-class evaluation protocol for Replica.

\noindent\textbf{Implementation Details.}
We run our system on a desktop equipped with an Intel i9-14900K CPU and an NVIDIA RTX 4090 GPU. Edge reasoning and node annotation generation are performed using qwen-vl-max \cite{qwen}, while qwen-turbo \cite{qwen} is employed to verify category consistency as described in Sec. \ref{subsec:VLM data engine}. During training, we utilize the Adam optimizer with an initial learning rate of \(1\times10^{-3}\), training the model for 600 epochs. The batch size is set to 256 for all datasets. The loss weights \([\lambda_{\text{hie}}, \lambda_{\text{reg}}, \lambda_{\text{aux}}]\) are set to \([0.5, 1.5, 1.5]\), and the dual-alignment balance factors \([\gamma_{\text{reg}}, \gamma_{\text{aux}}]\) are set to \([0.7, 0.7]\), based on preliminary hyperparameter tuning. Further implementation specifics, including the exact VLM prompts, are provided in the supplementary material.

\noindent\textbf{Metrics.}
We select open-vocabulary semantic segmentation as a representative task for evaluating our model's open-vocabulary scene understanding capabilities. Following the experimental setup in OpenScene \cite{openscene}, we use mIoU (\%) and mAcc (\%) as the evaluation metrics.

\subsection{Open-Vocabulary Scene Understanding}

\begin{table*}[t]
    \centering
    \caption{\textbf{Open-Vocabulary 3D Semantic Segmentation Results on ScanNetV2 and ScanNet200.}
    We report mIoU and mAcc. $^\dagger$ denotes numbers reported in the original paper (ScanNetV2) or reproduced by us (ScanNet200) based on LSeg.}
    \label{tab:scannet_v2_200_merged}
    
    \scriptsize
    \setlength{\tabcolsep}{4pt}
    \renewcommand{\arraystretch}{1.20}
    
    \begin{tabular}{c l cc cc}
    \toprule
    \textbf{Type} & \textbf{Method}
    & \multicolumn{2}{c}{\textbf{ScanNetV2}}
    & \multicolumn{2}{c}{\textbf{ScanNet200}} \\
    \cmidrule(lr){3-4}\cmidrule(lr){5-6}
    & & \textbf{mIoU} & \textbf{mAcc} & \textbf{mIoU} & \textbf{mAcc} \\
    \midrule
    
    \multirow{6}{*}{\makecell[c]{\textit{Closed}\\\textit{Vocabulary}}}
    & TangentConv~\cite{tangentconv}             & 40.9 & --   & --   & --   \\
    & TextureNet~\cite{texturenet}               & 54.8 & --   & --   & --   \\
    & ScanComplete~\cite{scancom}                & 56.6 & --   & --   & --   \\
    & DCM-Net~\cite{dcmnet}                      & 65.8 & --   & --   & --   \\
    & Mix3D~\cite{mix3d}                         & 73.6 & --   & --   & --   \\
    & MinkowskiNet~\cite{mink}                   & 69.2 & 77.7 & --   & --   \\
    \midrule
    
    \multirow{13}{*}{\makecell[c]{\textit{Open}\\\textit{Vocabulary}}}
    & MSeg Voting~\cite{msegvoting}              & 45.6 & 54.4 & --   & --   \\
    & PLA~\cite{pla}                             & 17.7 & 33.5 & 1.8  & 3.1  \\
    & RegionPLC~\cite{regionplc}                 & 43.8 & 65.6 & 6.5  & 15.9 \\
    & OpenMask3D~\cite{openmask3d}               & 34.0 & --   & 10.5 & --   \\
    & MaskClustering~\cite{maskclustering}       & 34.4 & 59.6 & \rd{11.8} & \rd{24.7} \\
    & OpenScene-2D$^\dagger$~\cite{openscene}    & 50.0 & 62.7 & 8.8  & 13.5 \\
    & OpenScene-3D$^\dagger$~\cite{openscene}    & 52.9 & 63.2 & 6.8  & 8.7  \\
    & OpenScene-2D3D$^\dagger$~\cite{openscene}  & 54.2 & 66.6 & 7.9  & 11.4 \\
    & DMA-text only~\cite{dma}                   & 50.5 & 63.7 & 8.7  & 11.3 \\
    & Mosaic3D~\cite{mosaic3d}                   & 48.9 & \fs{73.6} & \nd{12.4}  & \nd{25.1} \\
    & CUA-O3D~\cite{cuao3d}                      & \rd{55.3} & 65.6 & 6.4  & 8.8  \\
    & OV3D~\cite{ov3d}                           & \nd{57.3} & \rd{72.9} & 8.7  & --   \\
    & \textbf{RelGraphOV (Ours)}                 & \fs{58.4} & \nd{73.4} & \fs{14.5} & \fs{26.1} \\
    \bottomrule
    \end{tabular}

\end{table*}

\begin{table*}[t]
    \centering
    \caption{\textbf{Per-Category IoU on ScanNetV2}~\cite{scannet}. We report IoU (\%) for a representative subset of categories.}
    \label{tab:partial_category}
    
    \setlength{\tabcolsep}{4pt}
    \renewcommand{\arraystretch}{1.25}
    \newcolumntype{C}{>{\centering\arraybackslash}p{1.45cm}}
    
    \resizebox{\textwidth}{!}{
    \begin{tabular}{l|C|C|C|C|C|C|C|C|C|C}
        \toprule
         & cabinet & bed & door & window & fridge & curtain & shower curtain & toilet & bathtub & chair \\
        \midrule
        OpenScene~\cite{openscene}
         & \rd{43.6} & \rd{71.8} & \rd{49.8} & \rd{48.2} & \nd{49.4} & \rd{53.2} & \nd{0.0} & \rd{67.3} & \rd{61.4} & \rd{57.9} \\
        CUA-O3D~\cite{cuao3d}
         & \nd{44.7} & \nd{73.2} & \nd{50.5} & \nd{52.8} & \rd{39.6} & \nd{58.7} & \nd{0.0} & \nd{78.4} & \nd{62.3} & \fs{76.4} \\
        \textbf{Ours}
         & \fs{46.3} & \fs{76.8} & \fs{54.5} & \fs{58.3} & \fs{57.8} & \fs{66.6} & \fs{64.2} & \fs{83.6} & \fs{67.5} & \nd{71.7} \\
        \bottomrule
    \end{tabular}
    }
\end{table*}

As shown in Tab.~\ref{tab:scannet_v2_200_merged}, RelGraphOV achieves the best open-vocabulary mIoU on ScanNetV2 \cite{scannet}, competitive mAcc, and the best mIoU and mAcc on ScanNet200 \cite{scannet200} among the compared methods. These results suggest that relational context and the Adaptive Gated Dual-Stream Contextual GAT improve open-vocabulary feature refinement across both standard and long-tail label spaces.

\begin{figure*}[t!]
  \centering
  \includegraphics[width=0.98\textwidth]{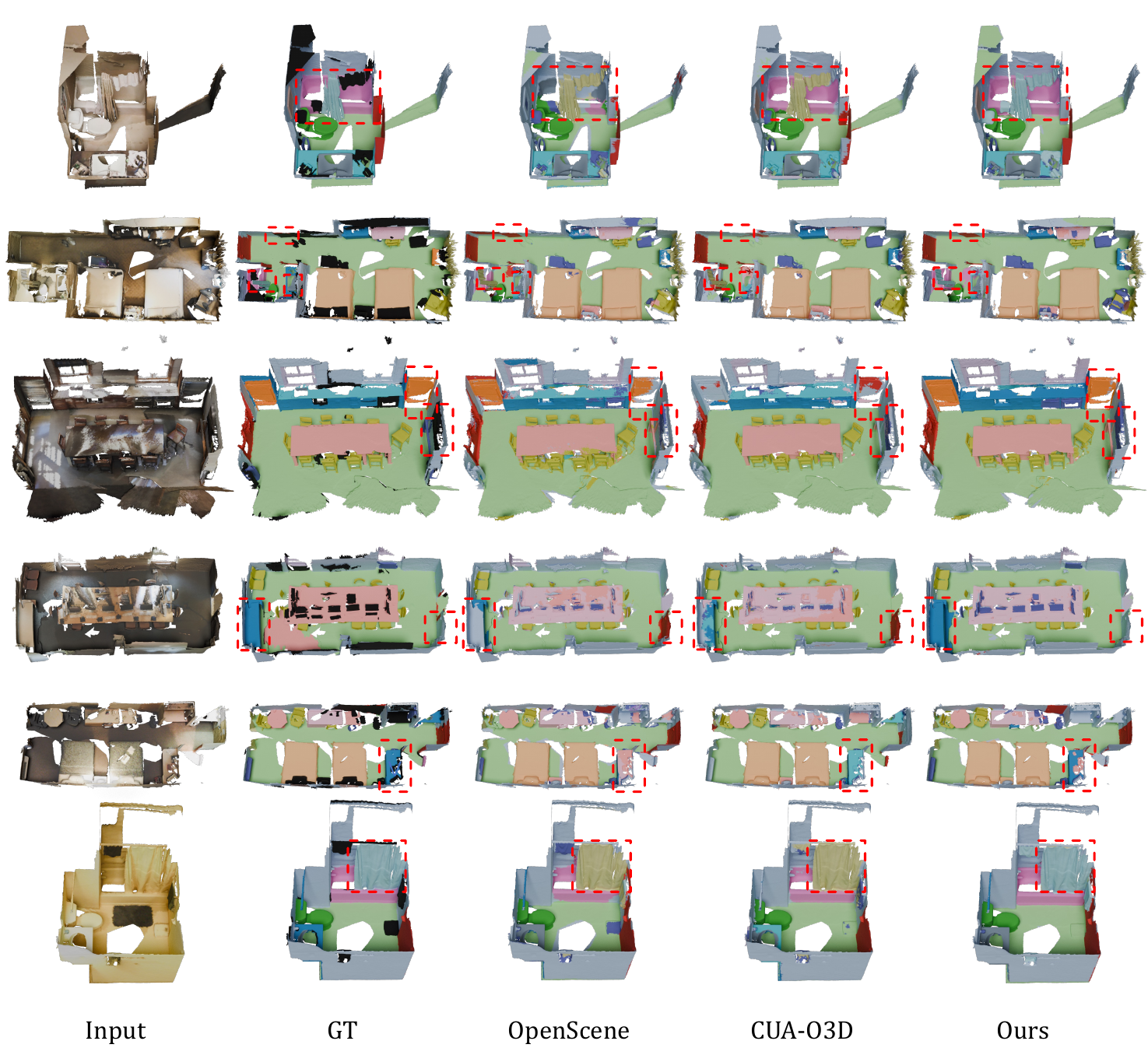}
  \caption{\textbf{Qualitative Comparisons.} Images of 3D semantic segmentation results on ScanNetV2 \cite{scannet}, including the example of a curtain and shower curtain as mentioned in the introduction. }
  \label{fig:more_compare}
\end{figure*}

On ScanNetV2 \cite{scannet}, our method improves over OV3D \cite{ov3d} in mIoU and over OpenScene-2D \cite{openscene} by +8.4\% mIoU and +10.7\% mAcc. The distinction between ``shower curtain'' and the general ``curtain'' category provides a representative case, as shown in Tab.~\ref{tab:partial_category} and Fig.~\ref{fig:more_compare}. OpenScene \cite{openscene} and CUA-O3D \cite{cuao3d} fail to recognize ``shower curtain'' in this setting, likely because they rely mainly on object appearance and group the points into the broader ``curtain'' class. In contrast, RelGraphOV uses VLM-reasoned neighborhood relations during message passing, improving the IoU for ``shower curtain'' to 64.2\%. 

    
    

We further evaluate the open-vocabulary capability and generalization performance of RelGraphOV on the challenging ScanNet200 \cite{scannet200} benchmark. This dataset contains 200 categories, ten times larger than commonly used previous benchmarks, posing severe long-tail challenges. Existing methods struggle here for two distinct reasons. Distillation-based approaches (e.g., OpenScene \cite{openscene}, DMA \cite{dma}) are constrained by the semantic coverage of their 2D teacher models. Conversely, methods directly using CLIP to extract instance features (e.g., MaskClustering \cite{maskclustering}, OpenMask3D \cite{openmask3d}) have an advantage on long-tail categories but often struggle on large, texture-less base objects like walls and floors due to incomplete multi-view geometric observations. On ScanNet200, RelGraphOV obtains 14.5\% mIoU and 26.1\% mAcc, outperforming the compared open-vocabulary baselines in both metrics. This result suggests that the dual-stream design helps combine the spatial robustness of dense LSeg features with the category generalization of CLIP features.

To further analyze the source of the improvement on ScanNet200, we split its categories into head, common, and tail groups. As shown in Tab.~\ref{tab:hct}, \textit{\method{}} achieves the best overall performance and shows clear advantages on common and tail categories. Although Mosaic3D performs better on head categories, our method is more effective on the more challenging common and tail groups, suggesting that relational context is particularly beneficial when local visual evidence is ambiguous or insufficient.

\begin{table}[t!]
\centering
\tiny
\setlength{\tabcolsep}{4.0pt}
\renewcommand{\arraystretch}{1.2} 
\caption{\textbf{Analysis on ScanNet200}~\cite{scannet200}. Following the official split, the 200 categories are divided into head, common, and tail groups. Each entry reports mIoU/mAcc (\%).}
\label{tab:hct}
\resizebox{0.86\linewidth}{!}{
\begin{tabular}{lcccc}
\toprule
Method & All & Head & Common & Tail \\
\midrule
OpenScene-2D~\cite{openscene}
& 8.8/13.5 & 19.6/29.2 & 3.7/8.4 & 1.5/2.4 \\
MaskClustering~\cite{maskclustering}
& \rd{11.8/24.7} & \rd{20.7/34.3} & \nd{8.6/20.6} & \nd{5.5/17.5} \\
Mosaic3D~\cite{mosaic3d}
& \nd{12.4/25.1} & \fs{28.9/43.1} & \rd{5.6/20.5} & \rd{1.3/9.8} \\
CUA-O3D\cite{cuao3d}
& 6.4/8.8 & 19.0/26.3 & 0.2/0.6 & 0.0/0.0 \\
\textbf{RelGraphOV}
& \fs{14.5/26.1} & \nd{26.1/36.3} & \fs{10.2/22.6} & \fs{6.3/18.1} \\
\bottomrule
\end{tabular}
}
\end{table}

Due to space limitations, additional qualitative comparisons and 3D open-vocabulary query results can be found in the supplementary material.

\subsection{Zero-Shot Cross-Dataset Evaluation}
To rigorously evaluate the cross-dataset generalization capabilities of our model, we conduct strict zero-shot inference on ScanNet$++$ \cite{scannetpp} and Replica \cite{replica}. Specifically, our model is trained exclusively on the real-world ScanNet dataset and evaluated directly on the target datasets without any prior training or fine-tuning on their data. 

\noindent\textbf{Evaluation on ScanNet++.}
We first evaluate on ScanNet++, a benchmark of higher-quality reconstructed indoor scenes, under its standard 100-class setting. As shown in Tab.~\ref{tab:scannetpp}, \textit{\method{}} clearly outperforms representative open-vocabulary 3D baselines, indicating that the learned relational feature refinement generalizes beyond the ScanNet validation setting.

\begin{table}[t]
\centering
\caption{\textbf{Zero-Shot Evaluation on ScanNet$++$}~\cite{scannetpp}. All methods are trained only on ScanNetV2 and evaluated directly on the 100-class ScanNet$++$ benchmark without fine-tuning.}
\label{tab:scannetpp}
\scriptsize
\setlength{\tabcolsep}{4pt}
\renewcommand{\arraystretch}{0.8}
\resizebox{\linewidth}{!}{
\begin{tabular}{lcccccc}
\toprule
Method & OpenScene-2D~\cite{openscene} & OpenScene-3D~\cite{openscene} & OpenScene-2D3D\cite{openscene} & CUA-O3D~\cite{cuao3d} & \textbf{RelGraphOV} \\
\midrule
mIoU & \nd{13.3} & 8.9  & \rd{11.7} & 9.3  & \fs{20.9} \\
mAcc & \nd{20.0} & 13.6 & \rd{15.3} & 13.7 & \fs{33.2} \\
\bottomrule
\end{tabular}
}
\end{table}

\begin{table*}[t]
\centering
\caption{\textbf{Zero-Shot Evaluation on Replica}~\cite{replica}. The Zero-Shot column indicates whether a method is evaluated without any training or fine-tuning on Replica. $^\dagger$~denotes results reproduced by us.}
\label{tab:replica_semseg}

\setlength{\tabcolsep}{6pt}
\renewcommand{\arraystretch}{1.15}

\resizebox{\textwidth}{!}{%
\begin{tabular}{l c cccccccc}
\toprule
\textbf{Method} & \textbf{Zero-Shot}
 & \multicolumn{2}{c}{\textit{All}} 
 & \multicolumn{2}{c}{\textit{Head}} 
 & \multicolumn{2}{c}{\textit{Common}} 
 & \multicolumn{2}{c}{\textit{Tail}} \\
\cmidrule(lr){3-4}\cmidrule(lr){5-6}\cmidrule(lr){7-8}\cmidrule(lr){9-10}
 &  & mIoU & mAcc & mIoU & mAcc & mIoU & mAcc & mIoU & mAcc \\
\midrule
OpenScene-2D$^\dagger$~\cite{openscene}
  & \cmark
  & 12.9 & 19.4
  & 31.8 & 43.1
  & 6.1  & 13.9
  & 1.2  & 1.2 \\
OpenScene-3D$^\dagger$~\cite{openscene}
  & \cmark
  & 10.4 & 17.4
  & 26.8 & 38.5
  & 4.5  & 13.5
  & 0.0  & 0.0 \\
OpenNeRF~\cite{opennerf} 
  & \xmark
  & \nd{20.4} & \nd{31.7} 
  & \nd{35.4} & \nd{46.2} 
  & \nd{20.1} & \nd{31.3} 
  & \nd{5.8}  & \fs{17.6} \\
\midrule
\textbf{RelGraphOV (Ours)}
  & \cmark
  & \fs{22.7} & \fs{32.7}
  & \fs{38.4} & \fs{55.2}
  & \fs{22.4} & \fs{33.7}
  & \fs{7.3}  & \nd{9.2} \\
\bottomrule
\end{tabular}%
}
\end{table*}

\noindent\textbf{Evaluation on Replica.}
Beyond the real-world scans of ScanNet++, we further assess generalization to the synthetic, photorealistic Replica dataset, which exhibits a larger domain gap from the ScanNet training data. As shown in Tab.~\ref{tab:replica_semseg}, RelGraphOV generalizes well to these unseen environments, achieving the highest overall mIoU (22.7\%) and mAcc (32.7\%) among the compared methods. The head/common/tail breakdown shows that our method improves rare-category mIoU, although OpenNeRF \cite{opennerf} obtains a higher tail mAcc. These results suggest that preserving CLIP-based semantic cues in the auxiliary stream helps cross-dataset generalization.

\subsection{Ablation Study \& Analysis}

\begin{table*}[t]
\centering
\setlength{\tabcolsep}{5pt}
\renewcommand{\arraystretch}{1.12}

\begin{minipage}[t]{0.485\textwidth}
    \centering
    \caption{\textbf{Ablation Study.} We validate the effectiveness of our relationship-aware 3D scene understanding framework and the dual-stream architecture.}
    \label{tab:ablation1}
    \resizebox{\textwidth}{!}{%
        \begin{tabular}{l c c c c}
            \toprule
            \textbf{Setting} &
            \multicolumn{2}{c}{\textbf{ScanNetV2\cite{scannet}}} &
            \multicolumn{2}{c}{\textbf{ScanNet200\cite{scannet200}}} \\
            & \textbf{mIoU} & \textbf{mAcc} & \textbf{mIoU} & \textbf{mAcc} \\
            \midrule
            w/o scene graph          & 50.7 & 62.6 & 9.5 & 14.4 \\
            w/o relation guidance    & 55.5 & 70.7 & 11.0 & 20.1 \\
            w/o global feature       & \rd{56.0} & \rd{71.3} & \rd{11.1} & \rd{20.2} \\
            w/o dual graph fusion    & \nd{57.9} & \nd{73.0} & \nd{12.7} & \nd{22.9} \\
            \textbf{Ours (full)}       & \fs{58.4} & \fs{73.4} & \fs{14.5} & \fs{26.1} \\
            \bottomrule
        \end{tabular}%
    }
\end{minipage}
\hfill
\begin{minipage}[t]{0.47\textwidth}
    \centering
    \caption{\textbf{Ablation Study.} We validate the effectiveness of our hierarchical contrastive learning and dual-alignment strategy.}
    \label{tab:ablation2}
    \resizebox{\textwidth}{!}{%
        \begin{tabular}{l c c c c}
            \toprule
            \textbf{Setting} &
            \multicolumn{2}{c}{\textbf{ScanNetV2\cite{scannet}}} &
            \multicolumn{2}{c}{\textbf{ScanNet200\cite{scannet200}}} \\
            & \textbf{mIoU} & \textbf{mAcc} & \textbf{mIoU} & \textbf{mAcc} \\
            \midrule
            w/o \(\mathcal{L}_{\text{hie}}\) & 52.2 & 65.5 & 7.8  & 12.4 \\
            w/o \(\mathcal{L}_{\text{reg}}\) & \rd{54.9} & \nd{71.7} & \rd{10.9} & \rd{20.1} \\
            w/o \(\mathcal{L}_{\text{aux}}\) & \nd{55.9} & \rd{70.4} & \nd{13.4} & \nd{25.6} \\
            \textbf{Ours (full)}               & \fs{58.4} & \fs{73.4} & \fs{14.5} & \fs{26.1} \\
            \bottomrule
        \end{tabular}%
    }
\end{minipage}
\end{table*}

\noindent\textbf{Effectiveness of Dual-Stream Architecture and Relation-guided GAT.}
To validate the structural components of our framework, we conduct a multi-step ablation study (Tab.~\ref{tab:ablation1}). First, to evaluate the role of message passing, we remove all edges, leaving only the initial node features (w/o scene graph). This degenerates our model into merely pooling LSeg features within isolated 3D object masks without any CLIP feature injection or inter-node communication, causing a clear performance drop and highlighting the need for contextual aggregation. Second, we verify the importance of edge-feature guidance by removing all edge features while preserving graph topology (w/o relation guidance). This reduction to a vanilla GAT leads to a noticeable drop, indicating that relationship semantics help filter geometric noise. Finally, to validate our dual-stream design, we evaluate a variant (w/o dual graph fusion) where LSeg and CLIP features are naively concatenated and processed in a single GNN stream. The resulting performance drop supports the ``Feature Interference'' hypothesis: mixing modalities in a single stream leads to cross-contamination. Our decoupled propagation and adaptive gated fusion are important for absorbing complementary semantics.

\noindent\textbf{Global Information Injection.}
An ablation of the global information extraction and injection module is performed, with results in Tab.~\ref{tab:ablation1} showing its contribution. Removing it (w/o global feature) causes a clear performance drop, indicating its role in enhancing nodes' global contextual awareness. The module integrates scene-level context to mitigate the limited receptive field of traditional GNNs and supports long-range dependency modeling in our architecture.

\noindent\textbf{Hierarchical Contrastive and Dual-Alignment Learning Strategy.}
We evaluate our proposed loss scheme for aligning node features with textual embeddings (Tab.~\ref{tab:ablation2}). First, removing the hierarchical contrastive loss (\(\mathcal{L}_{hie}\)) leads to a substantial performance drop, particularly on the challenging ScanNet200 benchmark. This indicates that \(\mathcal{L}_{hie}\) is important for maintaining category-level distinctions and reducing inter-class confusion. Furthermore, ablating the generalized dual-alignment losses (\(\mathcal{L}_{reg}\) and \(\mathcal{L}_{aux}\)) also leads to clear performance deterioration, highlighting the importance of preserving specific priors during feature refinement. Specifically, removing the main stream alignment (w/o \(\mathcal{L}_{reg}\)) degrades the LSeg prior and weakens segmentation boundaries. Similarly, removing the auxiliary stream alignment (w/o \(\mathcal{L}_{aux}\)) leads to semantic drift; without anchoring features to the initial CLIP space, the auxiliary stream loses open-vocabulary generalizability and fails to effectively supplement the main stream.

\section{Conclusion}
\label{sec:conclusion}

In this work, we presented \textit{\method{}}, a relationship-aware framework that incorporates environmental context into open-vocabulary 3D scene understanding via 3D scene graphs. We introduced a multi-view reasoning based scene graph construction pipeline to infer object relationships without manual annotations, and proposed an Adaptive Gated Dual-Stream Contextual GAT with hierarchical contrastive learning to enable robust contextual feature aggregation. Extensive experiments on real-world and synthetic benchmarks demonstrate state-of-the-art performance and strong cross-dataset generalization.

\noindent \textbf{Limitations.}
Our current implementation uses fixed LSeg and CLIP feature backbones to isolate the effect of relational contextual refinement. Recent dense vision-language backbones may provide stronger initial open-vocabulary features, and integrating them into the same relational refinement framework is a promising direction for future work. In addition, our current formulation focuses on static reconstructed scenes. Extending \textit{\method{}} to dynamic scenes with evolving objects and relationships requires temporally consistent scene-graph construction and update mechanisms.


\section*{Acknowledgments}
This work was supported by the National Key R\&D Program of China under Grant 2024YFB4505500 \& 2024YFB4505501. We also express our gratitude to all the anonymous reviewers for their professional and insightful comments.

%
%
\bibliographystyle{splncs04}
\bibliography{main}

@String(CVPR  = {IEEE Conf. Comput. Vis. Pattern Recog.})

@String(ICCV  = {Int. Conf. Comput. Vis.})

@String(ECCV  = {Eur. Conf. Comput. Vis.})

@String(NeurIPS = {Adv. Neural Inform. Process. Syst.})

@String(ICML  = {Int. Conf. Mach. Learn.})

@String(ICLR  = {Int. Conf. Learn. Represent.})

@String(CVPR  = {CVPR})

@String(ICCV  = {ICCV})

@String(ECCV  = {ECCV})

@String(NeurIPS = {NeurIPS})

@String(ICML  = {ICML})

@String(ICLR  = {ICLR})

@article{SoM,
  author       = {Jianwei Yang and
                  Hao Zhang and
                  Feng Li and
                  Xueyan Zou and
                  Chunyuan Li and
                  Jianfeng Gao},
  title        = {Set-of-Mark Prompting Unleashes Extraordinary Visual Grounding in
                  {GPT-4V}},
  journal      = {CoRR},
  volume       = {abs/2310.11441},
  year         = {2023},
  eprinttype    = {arXiv},
  eprint       = {2310.11441},
}

@inproceedings{mix3d,
  author       = {Alexey Nekrasov and
                  Jonas Schult and
                  Or Litany and
                  Bastian Leibe and
                  Francis Engelmann},
  title        = {Mix3D: Out-of-Context Data Augmentation for 3D Scenes},
  booktitle    = {International Conference on 3D Vision, 3DV 2021, London, United Kingdom,
                  December 1-3, 2021},
  pages        = {116--125},
  publisher    = {{IEEE}},
  year         = {2021},
}

@inproceedings{ov3d,
  author       = {Li Jiang and
                  Shaoshuai Shi and
                  Bernt Schiele},
  title        = {Open-Vocabulary 3D Semantic Segmentation with Foundation Models},
  booktitle    = {{IEEE/CVF} Conference on Computer Vision and Pattern Recognition,
                  {CVPR} 2024, Seattle, WA, USA, June 16-22, 2024},
  pages        = {21284--21294},
  publisher    = {{IEEE}},
  year         = {2024},
}

@inproceedings{mink,
  author       = {Christopher B. Choy and
                  JunYoung Gwak and
                  Silvio Savarese},
  title        = {4D Spatio-Temporal ConvNets: Minkowski Convolutional Neural Networks},
  booktitle    = {{IEEE} Conference on Computer Vision and Pattern Recognition, {CVPR}
                  2019, Long Beach, CA, USA, June 16-20, 2019},
  pages        = {3075--3084},
  publisher    = {Computer Vision Foundation / {IEEE}},
  year         = {2019},
}

@inproceedings{clip,
  author       = {Alec Radford and
                  Jong Wook Kim and
                  Chris Hallacy and
                  Aditya Ramesh and
                  Gabriel Goh and
                  Sandhini Agarwal and
                  Girish Sastry and
                  Amanda Askell and
                  Pamela Mishkin and
                  Jack Clark and
                  Gretchen Krueger and
                  Ilya Sutskever},
  editor       = {Marina Meila and
                  Tong Zhang},
  title        = {Learning Transferable Visual Models From Natural Language Supervision},
  booktitle    = {Proceedings of the 38th International Conference on Machine Learning,
                  {ICML} 2021, 18-24 July 2021, Virtual Event},
  series       = {Proceedings of Machine Learning Research},
  volume       = {139},
  pages        = {8748--8763},
  publisher    = {{PMLR}},
  year         = {2021},
}

@inproceedings{openscene,
  author       = {Songyou Peng and
                  Kyle Genova and
                  Chiyu Max Jiang and
                  Andrea Tagliasacchi and
                  Marc Pollefeys and
                  Thomas A. Funkhouser},
  title        = {OpenScene: 3D Scene Understanding with Open Vocabularies},
  booktitle    = {{IEEE/CVF} Conference on Computer Vision and Pattern Recognition,
                  {CVPR} 2023, Vancouver, BC, Canada, June 17-24, 2023},
  pages        = {815--824},
  publisher    = {{IEEE}},
  year         = {2023},
}

@inproceedings{cuao3d,
  author       = {Jinlong Li and
                  Cristiano Saltori and
                  Fabio Poiesi and
                  Nicu Sebe},
  title        = {Cross-Modal and Uncertainty-Aware Agglomeration for Open-Vocabulary
                  3D Scene Understanding},
  booktitle    = {{IEEE/CVF} Conference on Computer Vision and Pattern Recognition,
                  {CVPR} 2025, Nashville, TN, USA, June 11-15, 2025},
  pages        = {19390--19400},
  publisher    = {Computer Vision Foundation / {IEEE}},
  year         = {2025},
}

@inproceedings{openmask3d,
  author       = {Ay{\c{c}}a Takmaz and
                  Elisabetta Fedele and
                  Robert W. Sumner and
                  Marc Pollefeys and
                  Federico Tombari and
                  Francis Engelmann},
  editor       = {Alice Oh and
                  Tristan Naumann and
                  Amir Globerson and
                  Kate Saenko and
                  Moritz Hardt and
                  Sergey Levine},
  title        = {OpenMask3D: Open-Vocabulary 3D Instance Segmentation},
  booktitle    = {Advances in Neural Information Processing Systems 36: Annual Conference
                  on Neural Information Processing Systems 2023, NeurIPS 2023, New Orleans,
                  LA, USA, December 10 - 16, 2023},
  year         = {2023},
}

@inproceedings{mpec,
  author       = {Yan Wang and
                  Baoxiong Jia and
                  Ziyu Zhu and
                  Siyuan Huang},
  title        = {Masked Point-Entity Contrast for Open-Vocabulary 3D Scene Understanding},
  booktitle    = {{IEEE/CVF} Conference on Computer Vision and Pattern Recognition,
                  {CVPR} 2025, Nashville, TN, USA, June 11-15, 2025},
  pages        = {14125--14136},
  publisher    = {Computer Vision Foundation / {IEEE}},
  year         = {2025},
}

@inproceedings{cot,
  author       = {Jason Wei and
                  Xuezhi Wang and
                  Dale Schuurmans and
                  Maarten Bosma and
                  Brian Ichter and
                  Fei Xia and
                  Ed H. Chi and
                  Quoc V. Le and
                  Denny Zhou},
  editor       = {Sanmi Koyejo and
                  S. Mohamed and
                  A. Agarwal and
                  Danielle Belgrave and
                  K. Cho and
                  A. Oh},
  title        = {Chain-of-Thought Prompting Elicits Reasoning in Large Language Models},
  booktitle    = {Advances in Neural Information Processing Systems 35: Annual Conference
                  on Neural Information Processing Systems 2022, NeurIPS 2022, New Orleans,
                  LA, USA, November 28 - December 9, 2022},
  year         = {2022},
}

@inproceedings{scannet,
  author       = {Angela Dai and
                  Angel X. Chang and
                  Manolis Savva and
                  Maciej Halber and
                  Thomas A. Funkhouser and
                  Matthias Nie{\ss}ner},
  title        = {ScanNet: Richly-Annotated 3D Reconstructions of Indoor Scenes},
  booktitle    = {2017 {IEEE} Conference on Computer Vision and Pattern Recognition,
                  {CVPR} 2017, Honolulu, HI, USA, July 21-26, 2017},
  pages        = {2432--2443},
  publisher    = {{IEEE} Computer Society},
  year         = {2017},
}

@inproceedings{scannet200,
  author       = {D{\'{a}}vid Rozenberszki and
                  Or Litany and
                  Angela Dai},
  editor       = {Shai Avidan and
                  Gabriel J. Brostow and
                  Moustapha Ciss{\'{e}} and
                  Giovanni Maria Farinella and
                  Tal Hassner},
  title        = {Language-Grounded Indoor 3D Semantic Segmentation in the Wild},
  booktitle    = {Computer Vision - {ECCV} 2022 - 17th European Conference, Tel Aviv,
                  Israel, October 23-27, 2022, Proceedings, Part {XXXIII}},
  series       = {Lecture Notes in Computer Science},
  volume       = {13693},
  pages        = {125--141},
  publisher    = {Springer},
  year         = {2022},
}

@inproceedings{dino,
  author       = {Mathilde Caron and
                  Hugo Touvron and
                  Ishan Misra and
                  Herv{\'{e}} J{\'{e}}gou and
                  Julien Mairal and
                  Piotr Bojanowski and
                  Armand Joulin},
  title        = {Emerging Properties in Self-Supervised Vision Transformers},
  booktitle    = {2021 {IEEE/CVF} International Conference on Computer Vision, {ICCV}
                  2021, Montreal, QC, Canada, October 10-17, 2021},
  pages        = {9630--9640},
  publisher    = {{IEEE}},
  year         = {2021},
  bibsource    = {dblp computer science bibliography, https://dblp.org}
}

@article{dinov2,
  author       = {Maxime Oquab and
                  Timoth{\'{e}}e Darcet and
                  Th{\'{e}}o Moutakanni and
                  Huy V. Vo and
                  Marc Szafraniec and
                  Vasil Khalidov and
                  Pierre Fernandez and
                  Daniel Haziza and
                  Francisco Massa and
                  Alaaeldin El{-}Nouby and
                  Mido Assran and
                  Nicolas Ballas and
                  Wojciech Galuba and
                  Russell Howes and
                  Po{-}Yao Huang and
                  Shang{-}Wen Li and
                  Ishan Misra and
                  Michael Rabbat and
                  Vasu Sharma and
                  Gabriel Synnaeve and
                  Hu Xu and
                  Herv{\'{e}} J{\'{e}}gou and
                  Julien Mairal and
                  Patrick Labatut and
                  Armand Joulin and
                  Piotr Bojanowski},
  title        = {DINOv2: Learning Robust Visual Features without Supervision},
  journal      = {Trans. Mach. Learn. Res.},
  volume       = {2024},
  year         = {2024},
}

@inproceedings{sam,
  author       = {Alexander Kirillov and
                  Eric Mintun and
                  Nikhila Ravi and
                  Hanzi Mao and
                  Chlo{\'{e}} Rolland and
                  Laura Gustafson and
                  Tete Xiao and
                  Spencer Whitehead and
                  Alexander C. Berg and
                  Wan{-}Yen Lo and
                  Piotr Doll{\'{a}}r and
                  Ross B. Girshick},
  title        = {Segment Anything},
  booktitle    = {{IEEE/CVF} International Conference on Computer Vision, {ICCV} 2023,
                  Paris, France, October 1-6, 2023},
  pages        = {3992--4003},
  publisher    = {{IEEE}},
  year         = {2023},
}

@misc{sam2,
      title={SAM 2: Segment Anything in Images and Videos}, 
      author={Nikhila Ravi and Valentin Gabeur and Yuan-Ting Hu and Ronghang Hu and Chaitanya Ryali and Tengyu Ma and Haitham Khedr and Roman Rädle and Chloe Rolland and Laura Gustafson and Eric Mintun and Junting Pan and Kalyan Vasudev Alwala and Nicolas Carion and Chao-Yuan Wu and Ross Girshick and Piotr Dollár and Christoph Feichtenhofer},
      year={2024},
      eprint={2408.00714},
      archivePrefix={arXiv},
      primaryClass={cs.CV},
}

@inproceedings{maskattentionmask,
  author       = {Bowen Cheng and
                  Ishan Misra and
                  Alexander G. Schwing and
                  Alexander Kirillov and
                  Rohit Girdhar},
  title        = {Masked-attention Mask Transformer for Universal Image Segmentation},
  booktitle    = {{IEEE/CVF} Conference on Computer Vision and Pattern Recognition,
                  {CVPR} 2022, New Orleans, LA, USA, June 18-24, 2022},
  pages        = {1280--1289},
  publisher    = {{IEEE}},
  year         = {2022},
}

@inproceedings{oneformer,
  author       = {Jitesh Jain and
                  Jiachen Li and
                  MangTik Chiu and
                  Ali Hassani and
                  Nikita Orlov and
                  Humphrey Shi},
  title        = {OneFormer: One Transformer to Rule Universal Image Segmentation},
  booktitle    = {{IEEE/CVF} Conference on Computer Vision and Pattern Recognition,
                  {CVPR} 2023, Vancouver, BC, Canada, June 17-24, 2023},
  pages        = {2989--2998},
  publisher    = {{IEEE}},
  year         = {2023},
}

@inproceedings{seeaao,
  author       = {Xueyan Zou and
                  Jianwei Yang and
                  Hao Zhang and
                  Feng Li and
                  Linjie Li and
                  Jianfeng Wang and
                  Lijuan Wang and
                  Jianfeng Gao and
                  Yong Jae Lee},
  editor       = {Alice Oh and
                  Tristan Naumann and
                  Amir Globerson and
                  Kate Saenko and
                  Moritz Hardt and
                  Sergey Levine},
  title        = {Segment Everything Everywhere All at Once},
  booktitle    = {Advances in Neural Information Processing Systems 36: Annual Conference
                  on Neural Information Processing Systems 2023, NeurIPS 2023, New Orleans,
                  LA, USA, December 10 - 16, 2023},
  year         = {2023},
}

@inproceedings{groundingdino,
  author       = {Shilong Liu and
                  Zhaoyang Zeng and
                  Tianhe Ren and
                  Feng Li and
                  Hao Zhang and
                  Jie Yang and
                  Qing Jiang and
                  Chunyuan Li and
                  Jianwei Yang and
                  Hang Su and
                  Jun Zhu and
                  Lei Zhang},
  editor       = {Ales Leonardis and
                  Elisa Ricci and
                  Stefan Roth and
                  Olga Russakovsky and
                  Torsten Sattler and
                  G{\"{u}}l Varol},
  title        = {Grounding {DINO:} Marrying {DINO} with Grounded Pre-training for Open-Set
                  Object Detection},
  booktitle    = {Computer Vision - {ECCV} 2024 - 18th European Conference, Milan, Italy,
                  September 29-October 4, 2024, Proceedings, Part {XLVII}},
  series       = {Lecture Notes in Computer Science},
  volume       = {15105},
  pages        = {38--55},
  publisher    = {Springer},
  year         = {2024},
}

@inproceedings{osprey,
  author       = {Yuqian Yuan and
                  Wentong Li and
                  Jian Liu and
                  Dongqi Tang and
                  Xinjie Luo and
                  Chi Qin and
                  Lei Zhang and
                  Jianke Zhu},
  title        = {Osprey: Pixel Understanding with Visual Instruction Tuning},
  booktitle    = {{IEEE/CVF} Conference on Computer Vision and Pattern Recognition,
                  {CVPR} 2024, Seattle, WA, USA, June 16-22, 2024},
  pages        = {28202--28211},
  publisher    = {{IEEE}},
  year         = {2024},
}

@article{dam,
  author       = {Long Lian and
                  Yifan Ding and
                  Yunhao Ge and
                  Sifei Liu and
                  Hanzi Mao and
                  Boyi Li and
                  Marco Pavone and
                  Ming{-}Yu Liu and
                  Trevor Darrell and
                  Adam Yala and
                  Yin Cui},
  title        = {Describe Anything: Detailed Localized Image and Video Captioning},
  journal      = {CoRR},
  volume       = {abs/2504.16072},
  year         = {2025},
  eprinttype    = {arXiv},
  eprint       = {2504.16072},
}

@inproceedings{silc,
  author       = {Muhammad Ferjad Naeem and
                  Yongqin Xian and
                  Xiaohua Zhai and
                  Lukas Hoyer and
                  Luc Van Gool and
                  Federico Tombari},
  editor       = {Ales Leonardis and
                  Elisa Ricci and
                  Stefan Roth and
                  Olga Russakovsky and
                  Torsten Sattler and
                  G{\"{u}}l Varol},
  title        = {{SILC:} Improving Vision Language Pretraining with Self-distillation},
  booktitle    = {Computer Vision - {ECCV} 2024 - 18th European Conference, Milan, Italy,
                  September 29-October 4, 2024, Proceedings, Part {XXI}},
  series       = {Lecture Notes in Computer Science},
  volume       = {15079},
  pages        = {38--55},
  publisher    = {Springer},
  year         = {2024},
}

@inproceedings{catseg,
  author       = {Seokju Cho and
                  Heeseong Shin and
                  Sunghwan Hong and
                  Anurag Arnab and
                  Paul Hongsuck Seo and
                  Seungryong Kim},
  title        = {CAT-Seg: Cost Aggregation for Open-Vocabulary Semantic Segmentation},
  booktitle    = {{IEEE/CVF} Conference on Computer Vision and Pattern Recognition,
                  {CVPR} 2024, Seattle, WA, USA, June 16-22, 2024},
  pages        = {4113--4123},
  publisher    = {{IEEE}},
  year         = {2024},
}

@inproceedings{lseg,
  author       = {Boyi Li and
                  Kilian Q. Weinberger and
                  Serge J. Belongie and
                  Vladlen Koltun and
                  Ren{\'{e}} Ranftl},
  title        = {Language-driven Semantic Segmentation},
  booktitle    = {The Tenth International Conference on Learning Representations, {ICLR}
                  2022, Virtual Event, April 25-29, 2022},
  publisher    = {OpenReview.net},
  year         = {2022},
}

@inproceedings{dma,
  author       = {Ruihuang Li and
                  Zhengqiang Zhang and
                  Chenhang He and
                  Zhiyuan Ma and
                  Vishal M. Patel and
                  Lei Zhang},
  editor       = {Ales Leonardis and
                  Elisa Ricci and
                  Stefan Roth and
                  Olga Russakovsky and
                  Torsten Sattler and
                  G{\"{u}}l Varol},
  title        = {Dense Multimodal Alignment for Open-Vocabulary 3D Scene Understanding},
  booktitle    = {Computer Vision - {ECCV} 2024 - 18th European Conference, Milan, Italy,
                  September 29-October 4, 2024, Proceedings, Part {XLIX}},
  series       = {Lecture Notes in Computer Science},
  volume       = {15107},
  pages        = {416--434},
  publisher    = {Springer},
  year         = {2024},
}

@inproceedings{regionplc,
  author       = {Jihan Yang and
                  Runyu Ding and
                  Weipeng Deng and
                  Zhe Wang and
                  Xiaojuan Qi},
  title        = {RegionPLC: Regional Point-Language Contrastive Learning for Open-World
                  3D Scene Understanding},
  booktitle    = {{IEEE/CVF} Conference on Computer Vision and Pattern Recognition,
                  {CVPR} 2024, Seattle, WA, USA, June 16-22, 2024},
  pages        = {19823--19832},
  publisher    = {{IEEE}},
  year         = {2024},
}

@inproceedings{pla,
  author       = {Runyu Ding and
                  Jihan Yang and
                  Chuhui Xue and
                  Wenqing Zhang and
                  Song Bai and
                  Xiaojuan Qi},
  title        = {{PLA:} Language-Driven Open-Vocabulary 3D Scene Understanding},
  booktitle    = {{IEEE/CVF} Conference on Computer Vision and Pattern Recognition,
                  {CVPR} 2023, Vancouver, BC, Canada, June 17-24, 2023},
  pages        = {7010--7019},
  publisher    = {{IEEE}},
  year         = {2023},
}

@misc{qwen,
      title={Qwen2.5 Technical Report}, 
      author={Qwen and : and An Yang and Baosong Yang and Beichen Zhang and Binyuan Hui and Bo Zheng and Bowen Yu and Chengyuan Li and Dayiheng Liu and Fei Huang and Haoran Wei and Huan Lin and Jian Yang and Jianhong Tu and Jianwei Zhang and Jianxin Yang and Jiaxi Yang and Jingren Zhou and Junyang Lin and Kai Dang and Keming Lu and Keqin Bao and Kexin Yang and Le Yu and Mei Li and Mingfeng Xue and Pei Zhang and Qin Zhu and Rui Men and Runji Lin and Tianhao Li and Tianyi Tang and Tingyu Xia and Xingzhang Ren and Xuancheng Ren and Yang Fan and Yang Su and Yichang Zhang and Yu Wan and Yuqiong Liu and Zeyu Cui and Zhenru Zhang and Zihan Qiu},
      year={2025},
      eprint={2412.15115},
      archivePrefix={arXiv},
      primaryClass={cs.CL},
}

@inproceedings{tangentconv,
  author       = {Maxim Tatarchenko and
                  Jaesik Park and
                  Vladlen Koltun and
                  Qian{-}Yi Zhou},
  title        = {Tangent Convolutions for Dense Prediction in 3D},
  booktitle    = {2018 {IEEE} Conference on Computer Vision and Pattern Recognition,
                  {CVPR} 2018, Salt Lake City, UT, USA, June 18-22, 2018},
  pages        = {3887--3896},
  publisher    = {Computer Vision Foundation / {IEEE} Computer Society},
  year         = {2018},
}

@inproceedings{texturenet,
  author       = {Jingwei Huang and
                  Haotian Zhang and
                  Li Yi and
                  Thomas A. Funkhouser and
                  Matthias Nie{\ss}ner and
                  Leonidas J. Guibas},
  title        = {TextureNet: Consistent Local Parametrizations for Learning From High-Resolution
                  Signals on Meshes},
  booktitle    = {{IEEE} Conference on Computer Vision and Pattern Recognition, {CVPR}
                  2019, Long Beach, CA, USA, June 16-20, 2019},
  pages        = {4440--4449},
  publisher    = {Computer Vision Foundation / {IEEE}},
  year         = {2019},
}

@inproceedings{scancom,
  author       = {Angela Dai and
                  Daniel Ritchie and
                  Martin Bokeloh and
                  Scott Reed and
                  J{\"{u}}rgen Sturm and
                  Matthias Nie{\ss}ner},
  title        = {ScanComplete: Large-Scale Scene Completion and Semantic Segmentation
                  for 3D Scans},
  booktitle    = {2018 {IEEE} Conference on Computer Vision and Pattern Recognition,
                  {CVPR} 2018, Salt Lake City, UT, USA, June 18-22, 2018},
  pages        = {4578--4587},
  publisher    = {Computer Vision Foundation / {IEEE} Computer Society},
  year         = {2018},
}

@inproceedings{dcmnet,
  author       = {Jonas Schult and
                  Francis Engelmann and
                  Theodora Kontogianni and
                  Bastian Leibe},
  title        = {DualConvMesh-Net: Joint Geodesic and Euclidean Convolutions on 3D
                  Meshes},
  booktitle    = {2020 {IEEE/CVF} Conference on Computer Vision and Pattern Recognition,
                  {CVPR} 2020, Seattle, WA, USA, June 13-19, 2020},
  pages        = {8609--8619},
  publisher    = {Computer Vision Foundation / {IEEE}},
  year         = {2020},
}

@inproceedings{msegvoting,
  author       = {John Lambert and
                  Zhuang Liu and
                  Ozan Sener and
                  James Hays and
                  Vladlen Koltun},
  title        = {MSeg: {A} Composite Dataset for Multi-Domain Semantic Segmentation},
  booktitle    = {2020 {IEEE/CVF} Conference on Computer Vision and Pattern Recognition,
                  {CVPR} 2020, Seattle, WA, USA, June 13-19, 2020},
  pages        = {2876--2885},
  publisher    = {Computer Vision Foundation / {IEEE}},
  year         = {2020},
}

@inproceedings{maskclustering,
  author       = {Mi Yan and
                  Jiazhao Zhang and
                  Yan Zhu and
                  He Wang},
  title        = {MaskClustering: View Consensus Based Mask Graph Clustering for Open-Vocabulary
                  3D Instance Segmentation},
  booktitle    = {{IEEE/CVF} Conference on Computer Vision and Pattern Recognition,
                  {CVPR} 2024, Seattle, WA, USA, June 16-22, 2024},
  pages        = {28274--28284},
  publisher    = {{IEEE}},
  year         = {2024},
}

@inproceedings{mosaic3d,
  author       = {Junha Lee and
                  Chunghyun Park and
                  Jaesung Choe and
                  Yu{-}Chiang Frank Wang and
                  Jan Kautz and
                  Minsu Cho and
                  Christopher B. Choy},
  title        = {Mosaic3D: Foundation Dataset and Model for Open-Vocabulary 3D Segmentation},
  booktitle    = {{IEEE/CVF} Conference on Computer Vision and Pattern Recognition,
                  {CVPR} 2025, Nashville, TN, USA, June 11-15, 2025},
  pages        = {14089--14101},
  publisher    = {Computer Vision Foundation / {IEEE}},
  year         = {2025},
}

@inproceedings{opennerf,
  title     = {{OpenNeRF: Open Set 3D Neural Scene Segmentation with Pixel-Wise Features and Rendered Novel Views}},
  author    = {Engelmann, Francis and Manhardt, Fabian and Niemeyer, Michael and Tateno, Keisuke and Pollefeys, Marc and Tombari, Federico},
  booktitle = {International Conference on Learning Representations},
  year      = {2024}
}

@article{replica,
  title =   {The {R}eplica Dataset: A Digital Replica of Indoor Spaces},
  author =  {Julian Straub and Thomas Whelan and Lingni Ma and Yufan Chen and Erik Wijmans and Simon Green and Jakob J. Engel and Raul Mur-Artal and Carl Ren and Shobhit Verma and Anton Clarkson and Mingfei Yan and Brian Budge and Yajie Yan and Xiaqing Pan and June Yon and Yuyang Zou and Kimberly Leon and Nigel Carter and Jesus Briales and  Tyler Gillingham and  Elias Mueggler and Luis Pesqueira and Manolis Savva and Dhruv Batra and Hauke M. Strasdat and Renzo De Nardi and Michael Goesele and Steven Lovegrove and Richard Newcombe },
  journal = {arXiv preprint arXiv:1906.05797},
  year =    {2019}
}

@inproceedings{conceptgraph,
  author       = {Qiao Gu and
                  Ali Kuwajerwala and
                  Sacha Morin and
                  Krishna Murthy Jatavallabhula and
                  Bipasha Sen and
                  Aditya Agarwal and
                  Corban Rivera and
                  William Paul and
                  Kirsty Ellis and
                  Rama Chellappa and
                  Chuang Gan and
                  Celso Miguel de Melo and
                  Joshua B. Tenenbaum and
                  Antonio Torralba and
                  Florian Shkurti and
                  Liam Paull},
  title        = {ConceptGraphs: Open-Vocabulary 3D Scene Graphs for Perception and
                  Planning},
  booktitle    = {{IEEE} International Conference on Robotics and Automation, {ICRA}
                  2024, Yokohama, Japan, May 13-17, 2024},
  pages        = {5021--5028},
  publisher    = {{IEEE}},
  year         = {2024},
}

@inproceedings{openseg,
  author       = {Golnaz Ghiasi and
                  Xiuye Gu and
                  Yin Cui and
                  Tsung{-}Yi Lin},
  editor       = {Shai Avidan and
                  Gabriel J. Brostow and
                  Moustapha Ciss{\'{e}} and
                  Giovanni Maria Farinella and
                  Tal Hassner},
  title        = {Scaling Open-Vocabulary Image Segmentation with Image-Level Labels},
  booktitle    = {Computer Vision - {ECCV} 2022 - 17th European Conference, Tel Aviv,
                  Israel, October 23-27, 2022, Proceedings, Part {XXXVI}},
  series       = {Lecture Notes in Computer Science},
  volume       = {13696},
  pages        = {540--557},
  publisher    = {Springer},
  year         = {2022},
}

@inproceedings{wald2020learning,
  author       = {Johanna Wald and
                  Helisa Dhamo and
                  Nassir Navab and
                  Federico Tombari},
  title        = {Learning 3D Semantic Scene Graphs From 3D Indoor Reconstructions},
  booktitle    = {2020 {IEEE/CVF} Conference on Computer Vision and Pattern Recognition,
                  {CVPR} 2020, Seattle, WA, USA, June 13-19, 2020},
  pages        = {3960--3969},
  publisher    = {Computer Vision Foundation / {IEEE}},
  year         = {2020},
}

@inproceedings{scenegraphfusion,
  author       = {Shun{-}Cheng Wu and
                  Johanna Wald and
                  Keisuke Tateno and
                  Nassir Navab and
                  Federico Tombari},
  title        = {SceneGraphFusion: Incremental 3D Scene Graph Prediction From {RGB-D}
                  Sequences},
  booktitle    = {{IEEE} Conference on Computer Vision and Pattern Recognition, {CVPR}
                  2021, virtual, June 19-25, 2021},
  pages        = {7515--7525},
  publisher    = {Computer Vision Foundation / {IEEE}},
  year         = {2021},
}

@inproceedings{vlsat,
  author       = {Ziqin Wang and
                  Bowen Cheng and
                  Lichen Zhao and
                  Dong Xu and
                  Yang Tang and
                  Lu Sheng},
  title        = {{VL-SAT:} Visual-Linguistic Semantics Assisted Training for 3D Semantic
                  Scene Graph Prediction in Point Cloud},
  booktitle    = {{IEEE/CVF} Conference on Computer Vision and Pattern Recognition,
                  {CVPR} 2023, Vancouver, BC, Canada, June 17-24, 2023},
  pages        = {21560--21569},
  publisher    = {{IEEE}},
  year         = {2023},
}

@inproceedings{lang3dsg,
  author       = {Sebastian Koch and
                  Pedro Hermosilla and
                  Narunas Vaskevicius and
                  Mirco Colosi and
                  Timo Ropinski},
  title        = {Lang3DSG: Language-based contrastive pre-training for 3D Scene Graph
                  prediction},
  booktitle    = {International Conference on 3D Vision, 3DV 2024, Davos, Switzerland,
                  March 18-21, 2024},
  pages        = {1037--1047},
  publisher    = {{IEEE}},
  year         = {2024},
}

@inproceedings{open3dsg,
  author       = {Sebastian Koch and
                  Narunas Vaskevicius and
                  Mirco Colosi and
                  Pedro Hermosilla and
                  Timo Ropinski},
  title        = {Open3DSG: Open-Vocabulary 3D Scene Graphs from Point Clouds with Queryable
                  Objects and Open-Set Relationships},
  booktitle    = {{IEEE/CVF} Conference on Computer Vision and Pattern Recognition,
                  {CVPR} 2024, Seattle, WA, USA, June 16-22, 2024},
  pages        = {14183--14193},
  publisher    = {{IEEE}},
  year         = {2024},
}

@inproceedings{contextaware3dgrounding,
  author       = {Haonan Chang and
                  Kowndinya Boyalakuntla and
                  Shiyang Lu and
                  Siwei Cai and
                  Eric Pu Jing and
                  Shreesh Keskar and
                  Shijie Geng and
                  Adeeb Abbas and
                  Lifeng Zhou and
                  Kostas E. Bekris and
                  Abdeslam Boularias},
  editor       = {Jie Tan and
                  Marc Toussaint and
                  Kourosh Darvish},
  title        = {Context-Aware Entity Grounding with Open-Vocabulary 3D Scene Graphs},
  booktitle    = {Conference on Robot Learning, CoRL 2023, 6-9 November 2023, Atlanta,
                  GA, {USA}},
  series       = {Proceedings of Machine Learning Research},
  volume       = {229},
  pages        = {1950--1974},
  publisher    = {{PMLR}},
  year         = {2023},
}

@inproceedings{relation3d,
  author       = {Jiahao Lu and
                  Jiacheng Deng},
  title        = {Relation3D : Enhancing Relation Modeling for Point Cloud Instance
                  Segmentation},
  booktitle    = {{IEEE/CVF} Conference on Computer Vision and Pattern Recognition,
                  {CVPR} 2025, Nashville, TN, USA, June 11-15, 2025},
  pages        = {8889--8899},
  publisher    = {Computer Vision Foundation / {IEEE}},
  year         = {2025},
}

@inproceedings{relationfield,
  author       = {Sebastian Koch and
                  Johanna Wald and
                  Mirco Colosi and
                  Narunas Vaskevicius and
                  Pedro Hermosilla and
                  Federico Tombari and
                  Timo Ropinski},
  title        = {RelationField: Relate Anything in Radiance Fields},
  booktitle    = {{IEEE/CVF} Conference on Computer Vision and Pattern Recognition,
                  {CVPR} 2025, Nashville, TN, USA, June 11-15, 2025},
  pages        = {21706--21716},
  publisher    = {Computer Vision Foundation / {IEEE}},
  year         = {2025},
}

@inproceedings{scannetpp,
  author       = {Chandan Yeshwanth and
                  Yueh{-}Cheng Liu and
                  Matthias Nie{\ss}ner and
                  Angela Dai},
  title        = {ScanNet++: {A} High-Fidelity Dataset of 3D Indoor Scenes},
  booktitle    = {{IEEE/CVF} International Conference on Computer Vision, {ICCV} 2023,
                  Paris, France, October 1-6, 2023},
  pages        = {12--22},
  publisher    = {{IEEE}},
  year         = {2023},
}

@article{orbslam3,
  author       = {Carlos Campos and
                  Richard Elvira and
                  Juan J. G{\'{o}}mez Rodr{\'{\i}}guez and
                  Jos{\'{e}} M. M. Montiel and
                  Juan D. Tard{\'{o}}s},
  title        = {{ORB-SLAM3:} An Accurate Open-Source Library for Visual, Visual-Inertial,
                  and Multimap {SLAM}},
  journal      = {{IEEE} Trans. Robotics},
  volume       = {37},
  number       = {6},
  pages        = {1874--1890},
  year         = {2021},
  bibsource    = {dblp computer science bibliography, https://dblp.org}
}

@inproceedings{vggt,
  author    = {Wang, Jianyuan and Chen, Minghao and Karaev, Nikita and Vedaldi, Andrea and Rupprecht, Christian and Novotny, David},
  title     = {{VGGT}: Visual Geometry Grounded Transformer},
  booktitle = {Proceedings of the IEEE/CVF Conference on Computer Vision and Pattern Recognition (CVPR)},
  year      = {2025},
}

@inproceedings{cgslam,
  author       = {Jiarui Hu and
                  Xianhao Chen and
                  Boyin Feng and
                  Guanglin Li and
                  Liangjing Yang and
                  Hujun Bao and
                  Guofeng Zhang and
                  Zhaopeng Cui},
  editor       = {Ales Leonardis and
                  Elisa Ricci and
                  Stefan Roth and
                  Olga Russakovsky and
                  Torsten Sattler and
                  G{\"{u}}l Varol},
  title        = {{CG-SLAM:} Efficient Dense {RGB-D} {SLAM} in a Consistent Uncertainty-Aware
                  3D Gaussian Field},
  booktitle    = {Computer Vision - {ECCV} 2024 - 18th European Conference, Milan, Italy,
                  September 29-October 4, 2024, Proceedings, Part {XXV}},
  series       = {Lecture Notes in Computer Science},
  volume       = {15083},
  pages        = {93--112},
  publisher    = {Springer},
  year         = {2024},
  bibsource    = {dblp computer science bibliography, https://dblp.org}
}

@inproceedings{cpslam,
  author       = {Jiarui Hu and
                  Mao Mao and
                  Hujun Bao and
                  Guofeng Zhang and
                  Zhaopeng Cui},
  editor       = {Alice Oh and
                  Tristan Naumann and
                  Amir Globerson and
                  Kate Saenko and
                  Moritz Hardt and
                  Sergey Levine},
  title        = {{CP-SLAM:} Collaborative Neural Point-based {SLAM} System},
  booktitle    = {Advances in Neural Information Processing Systems 36: Annual Conference
                  on Neural Information Processing Systems 2023, NeurIPS 2023, New Orleans,
                  LA, USA, December 10 - 16, 2023},
  year         = {2023},
  bibsource    = {dblp computer science bibliography, https://dblp.org}
}

@article{atlasgs,
  title={AtlasGS: Atlanta-world guided surface reconstruction with implicit structured gaussians},
  author={Zhang, Xiyu and Bao, Chong and Chen, Yipeng and Zhai, Hongjia and Dong, Yitong and Bao, Hujun and Cui, Zhaopeng and Zhang, Guofeng},
  journal={Advances in Neural Information Processing Systems},
  volume={38},
  pages={15556--15580},
  year={2026}
}

\end{document}